\documentclass[conference]{IEEEtran}
\IEEEoverridecommandlockouts
\usepackage{cite}
\usepackage{amsmath,amssymb,amsfonts}
\usepackage{algorithmic}
\usepackage{graphicx}
\usepackage{textcomp}
\usepackage{xcolor}
\usepackage{textgreek}
\usepackage{subfigure}
\usepackage{booktabs}        
\usepackage{multirow}        
\usepackage{threeparttable}  

\def\BibTeX{{\rm B\kern-.05em{\sc i\kern-.025em b}\kern-.08em
    T\kern-.1667em\lower.7ex\hbox{E}\kern-.125emX}}
\begin{document}
\title{An Event-Driven E-Skin System with Dynamic Binary Scanning and real time SNN Classification\thanks{This work was supported by the Research Grants Council of
the HK SAR, China (Project No. CityU 11212823 )}}
\author{\IEEEauthorblockN{Gaishan Li\textsuperscript{†}, Zhengnan Fu\textsuperscript{†}, Anubhab Tripathi, Junyi Yang, Arindam Basu*}
\IEEEauthorblockA{\textit{Department of Electrical Engineering} \\
\textit{City University of Hongkong}\\
Hongkong SAR, China \\
Email: [gaishali-c, zhengnafu2-c, junyiyang8-c]@my.cityu.edu.hk, [atripath,arinbasu*]@cityu.edu.hk}}

\maketitle
\begin{abstract}
This paper presents a novel hardware system for high-speed, event-sparse sampling-based electronic skin (e-skin) that integrates sensing and neuromorphic computing. The system is built around a 16×16 piezoresistive tactile array with front-end and introduces a event-based binary scan search strategy to classify the digits. This event-driven strategy achieves a 12.8× reduction in scan counts, a 38.2× data compression rate and  a 28.4× equivalent dynamic range, a 99\% data sparsity, drastically reducing the data acquisition overhead. The resulting sparse data stream is processed by a multi-layer convolutional spiking neural network (Conv-SNN) implemented on an FPGA, which requires only 65\% of the computation and 15.6\% of the weight storage relative to a CNN. Despite these significant efficiency gains, the system maintains a high classification accuracy of 92.11\% for real-time handwritten digit recognition. Furthermore, a real neuromorphic tactile dataset using Address Event Representation (AER) is constructed. This work demonstrates a fully integrated, event-driven pipeline from analog sensing to neuromorphic classification, offering an efficient solution for robotic perception and human-computer interaction.

\end{abstract}


\section{Introduction}
Tactile sense is increasingly pivotal for enabling machines to interact physically and intelligently with their environment, finding critical applications in robotics for dexterous manipulation and in human-computer interaction for rich, physical interfaces [1], [2]. Among the various sensing technologies, resistive-based tactile systems offer significant advantages in terms of scalability and cost-effectiveness, facilitating their deployment over large areas [3]. These systems typically consist of arrays of sensor elements where applied pressure causes a measurable change in resistance, which is then digitized through scanning circuits and an analog front-end (AFE).

However, conventional frame-based scanning methodologies—which drive most current systems—suffer from fundamental efficiency bottlenecks [4], [5]. Inspired by imaging sensors, these approaches periodically poll all sensors in the array to construct a full "tactile frame" [6], yet they are inherently mismatched to the sparse spatiotemporal nature of tactile data. By continuously acquiring data from every sensor regardless of activity, such systems generate substantial redundancy, leading to excessive power consumption and high transmission overhead [7]. Although some studies have employed more complex algorithms to achieve contact timing detection and high-quality tactile image reconstruction, the practical deployment of these solutions in edge electronic skin systems remains challenging [8].

The field of dynamic tactile sensing and e-skin continues to face fundamental challenges beyond data acquisition: while event-driven scanning reduces data at the source, most systems still rely on conventional artificial neural networks (ANN) for classification[9,10]. This creates a critical mismatch: the sparse, event-based data must be reformatted into dense frames for ANN processing, undermining the efficiency gains from dynamic scanning[11]. ANNs further suffer from substantial computational demands and parameter counts that are ill-suited for embedded systems. In contrast, spiking neural networks (SNN) offer a more suitable alternative, maintaining the event-based nature throughout the processing pipeline while achieving higher compression rates and requiring fewer parameters [12]. This end-to-end neuromorphic approach, combining event-driven acquisition with SNN processing, remains largely unexplored in current e-skin implementations.

\begin{figure*}[t]
    \centering
    \includegraphics[width=0.9\linewidth]{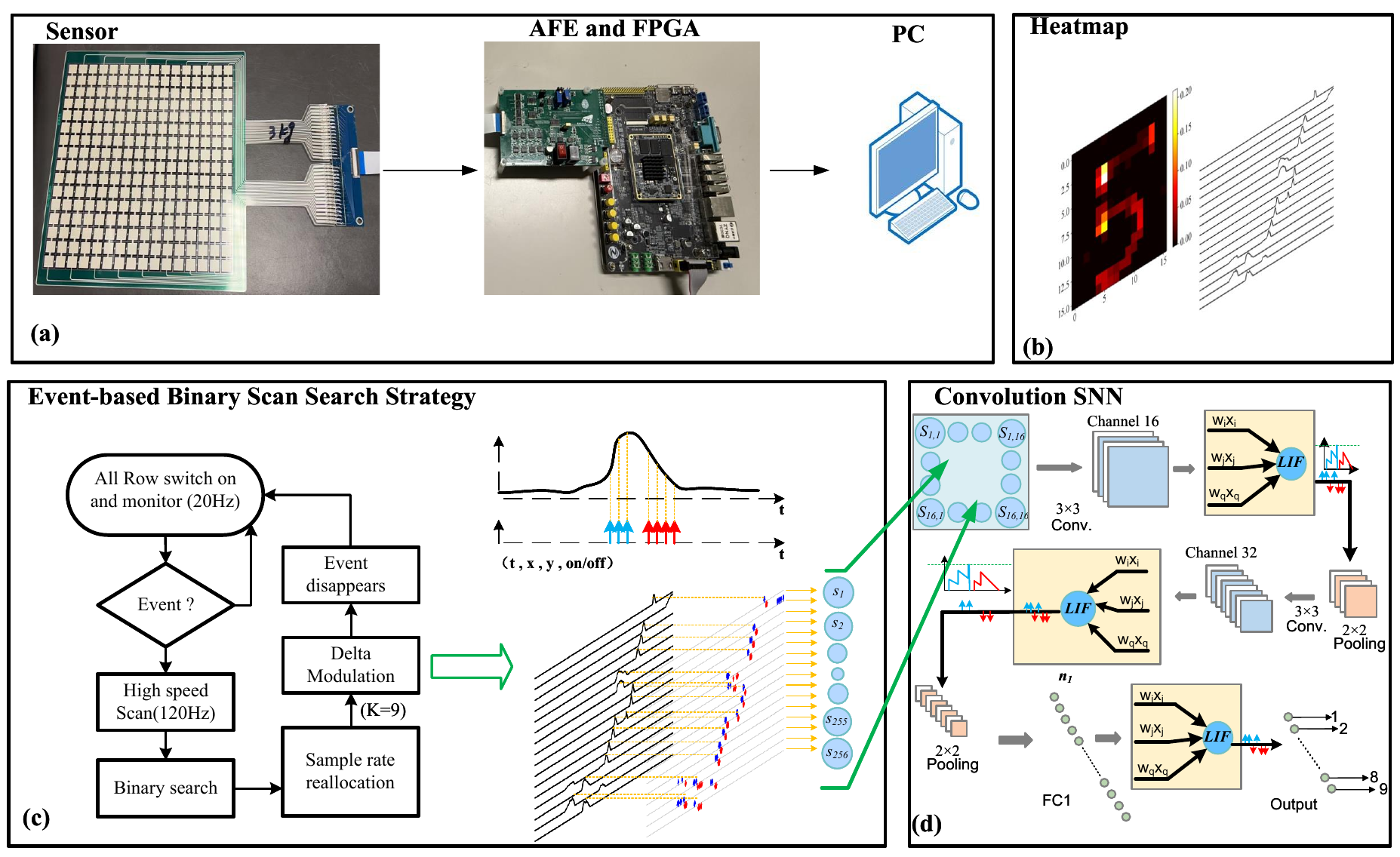}
    \caption{(a) Proposed overall neuromorphic hardware system, (b) Data heatmap, (c) Event-based Binary Scan Search Strategy (Pulse display of the 16 strongest channels), (d) Convolutional spiking neural network structure used in this system}
    \label{fig_all}
\end{figure*}
Our work addresses these limitations through a bio-inspired neuromorphic system that co-designs an event-driven scan strategy with efficient SNN processing. As illustrated in Fig. 1(a), the system comprises a sensor array, AFE, FPGA, and PC, forming a complete tactile data acquisition and processing pipeline. Tactile events are usually sparse and short-lived and the heat map of a tactile event is shown in Fig. \ref{fig_all}(b). We develop the event-based binary scan strategy in Fig. \ref{fig_all}(c) efficiently locates these active regions and results sparse, dynamic tactile signals are converted into spikes via delta modulation. The Conv-SNN architecture implemented on the FPGA (Fig. \ref{fig_all}(d)) processes these asynchronous event streams to efficiently extract temporal patterns with low neuronal activity, aligning perfectly with the sparse nature of the tactile data.

The principal innovations of this work are following: (1) a novel event-driven binary scan search strategy that increases data sparsity by 7.6×, a 12.8× reduction in scan counts and a 38.4× data compression rate \((\Delta=6\)); (2) a Conv-SNN on FPGA which requires only 65\% of the computation and 15.6\% of the weight storage of a conventional CNN while delivering a 92.11\% classification accuracy after quantilization (5bit); and (3) the establishment of a complete event-driven tactile processing pipeline, supported by a real neuromorphic AER dataset, which integrates analog sensing with neuromorphic computing.
\textbf{This work is also delivered to live demonstration track.}
\section{Sensor, front-end  and scanning algorithm}
\subsection{Sensor and front-end hardware system}
The distributed flexible pressure sensor array is fabricated via precision printing of nanomaterial-based force-sensitive layers and silver interconnects on a thin-film substrate shown in Fig. \ref{fig_all}(a). Configured as a 16×16 array within a 150 mm × 150 mm area, the sensor contains 256 units, each measuring 7.5 mm × 7.5 mm. The sensor exhibits piezoresistive behavior, with resistance decreasing as applied pressure increases. Calibration using a robotic arm confirms that resistance follows an inverse relationship with pressure, varying consistently from several GΩ down to 3 kΩ across the tested range. This well-defined characteristic provides key guidance for front-end amplifier design, ensuring appropriate dynamic range and readout resolution.

\begin{figure}[t]
    \centering
    \includegraphics[width=1\linewidth]{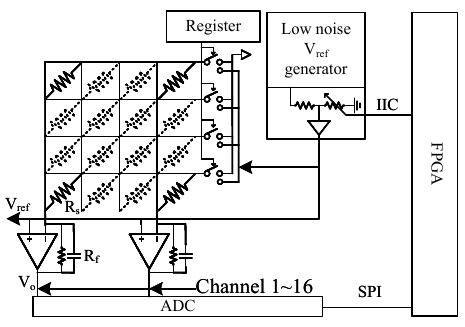}
    \caption{Schematic of the readout hardware system of tactile sensors}
    \label{fig_structure}
\end{figure}

Fig. \ref{fig_structure} illustrates a multi-channel data acquisition system built around an FPGA and a 16×16 resistive sensor array. Each sensor element is integrated into a parallel negative feedback circuit. The system employs switch chips (ADG734) controlled by a register (74HC16D) to connect array rows either to a precision reference voltage $V_{ref}$ or to GND. Selecting a row by connecting it to GND activates a feedback amplifier configuration using OPA2991 operational amplifiers. This switchable architecture enables multiple monitoring modes, enhancing the system's data compression capability. The output voltage is defined by $V_o = V_{ref}(1 + R_f / R_s)$, where the feedback resistor $R_f$ = 100 kΩ and the sensor resistance $R_s$ spans from several kΩ to hundreds of GΩ. To ensure stability against parasitic capacitance effects from the high-impedance source, a compensation capacitor $C_f$ is implemented across $R_f$ for Miller compensation.

The signal chain utilizes eight low-noise, rail-to-rail OPA2991 op-amps to fully exploit the ADC's dynamic range. A high-stability REF5025 voltage reference, with a temperature coefficient of 3 ppm/°C, provides the reference $V_{ref}$. The amplified analog signals are digitized by a 16-channel ADS7961 ADC capable of sampling at 1 MHz, with a programmable input range of 0 to $V_{ref}$ or 0 to $V_{ref}/2$. The digitized data is subsequently transmitted to the FPGA via the SPI communication protocol for further processing.

\subsection{Event-based Binary Scan Search Strategy}
As the integration density of tactile sensors increases, few tactile arrays with over a thousand sensing elements achieve sampling rates exceeding 100 Hz[13]. For the traditional one-by-one scanning method, the average number of scans is \(\frac{N}{2} \) which restricts the application of high spatial resolution tactile sensors. However, the spatial sparsity and temporal redundancy inherent in tactile signals provide potential solutions to this issue. Compressed sensing can exploit the spatial sparsity of tactile data to reduce the number of scans required for a complete readout[14]. Event-based encoding can exploit the temporal redundancy of tactile data to reduce the memory required for dataset storage[15]. However, this encoding method is not directly compatible with subsequent ANNs. In the context of tactile applications, our research proposes an event-based binary scan search strategy that leverages the characteristics of sensor arrays and circuit design. 

The flowchart of the proposed event-based binary scan search is shown in Fig. \ref{fig_all}(c), which illustrate this method using a piezoresistive crossbar (the same logic applies to capacitive arrays). In a crossbar, each row/column shares a data line, reducing wiring. The equivalent matrix, its scan controller, and readout are depicted in Fig. \ref{fig_structure}. When all rows are enabled, elements in a column form a parallel network sensed by the frontend; unpressed taxels are on the order of GΩ, while pressed taxels drop to kΩ, so the column’s parallel resistance reveals whether any activation is present. By configuring row switches, arbitrary subregions can be paralleled and tested in a single scan; once a column is flagged, a binary-search-style procedure rapidly excludes inactive subregions and localizes the touch. For our case that N = 256, the number of scans based on proposed strategy is reduced by about 12.8× and 1.6× compared with the traditional strategy [13] and the row and column switching strategy [16,17]. This work is applicable to all resistive and capacitive sensors with any crossbar connection configuration. After determining the touch hotspot, the sampling redistribution strategy continuously samples the 3 × 3 area around that point and continuously refocuses it on the tactile unit with the maximum pressure. Our method can enhance the dynamic range by 28.4×. 

\subsection{Dataset collection}
The work recruited 13 participants, each of whom repeatedly wrote digits from 1 to 9. Signals were acquired using a custom end-to-end hardware platform and transmitted to a computer for processing, resulting in a total of 760 samples with balanced class distribution. Each sample was collected by instructing participants to handwrite a digit within a 2s window, while raw ADC data were recorded at a sampling rate of 120 Hz. The acquired data were subsequently transferred to a computer for further processing and algorithmic validation. Some studies[17] leverage compressed sensing strategies to exploit the spatial sparsity of tactile data, reducing the number of scans required for a complete matrix readout. 
\section{Result and iscussion}
\subsection{Data compression results}
\begin{figure}[htbp]
    \centering
    \includegraphics[width=0.9\linewidth]{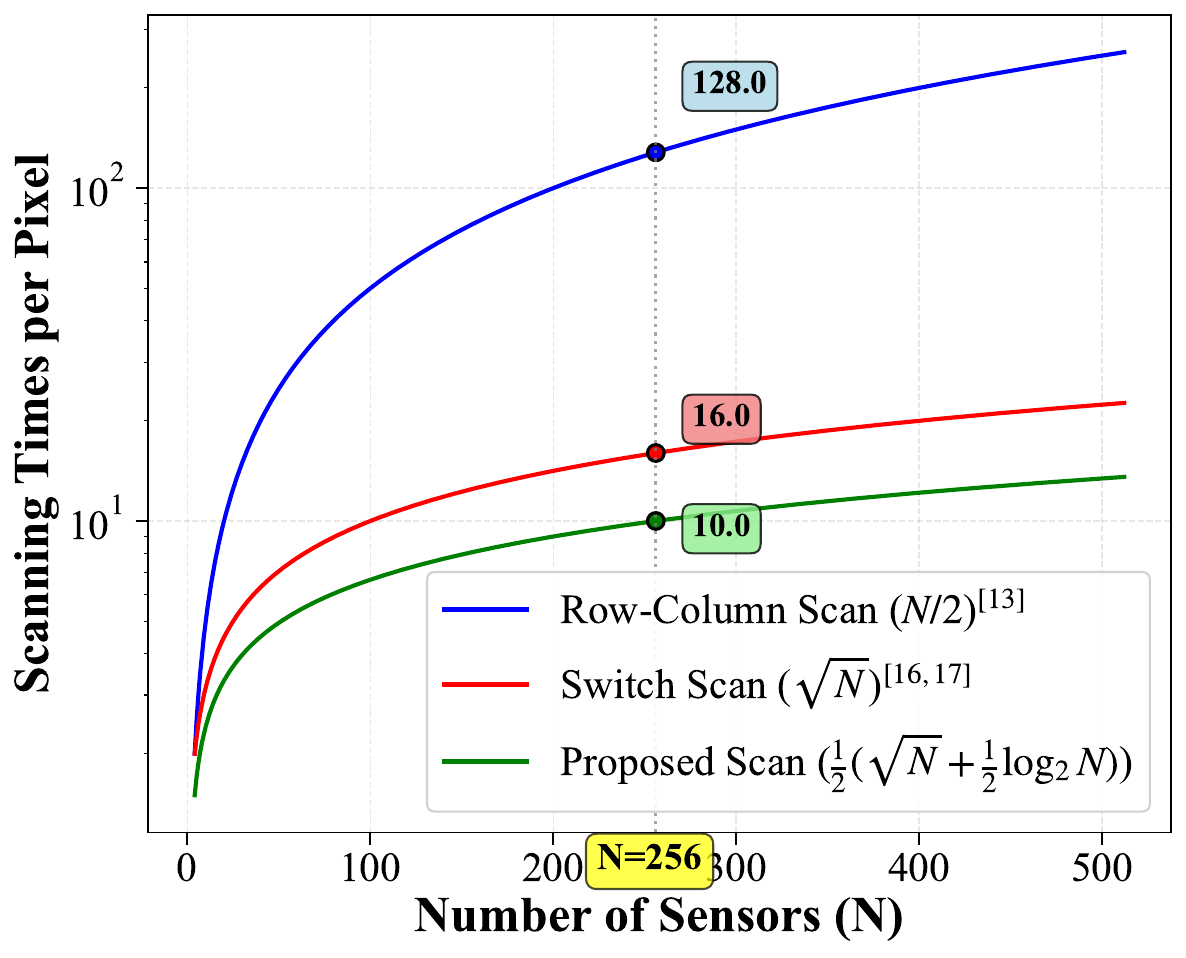}
    \caption{Comparison of traditional and proposed Scanning Strategies}
    \label{fig_scan}
\end{figure}

The proposed scan strategy enables the identification of activated tactile elements at any position with an average number of scans of \( \frac{1}{2} \times \left ( \sqrt{N} + \log_{2}{N} \right ) \), efficiently allocating ADC sampling bandwidth to localized regions. Compared to the methods of row-colunm scan [13] and switch scan [16],[17], our strategy reduces the scan times by 12.8×/1.6× when the number of sensors (N) is 256 as shown in Fig. \ref{fig_scan}. After the reallocation of sampling resources, the acquired data are converted into pulse sequences via delta modulation based on the spatial sparsity of tactile data. 

\begin{figure}[htbp]
    \centering
    \begin{subfigure}
        \centering
        \includegraphics[width=0.8\linewidth]{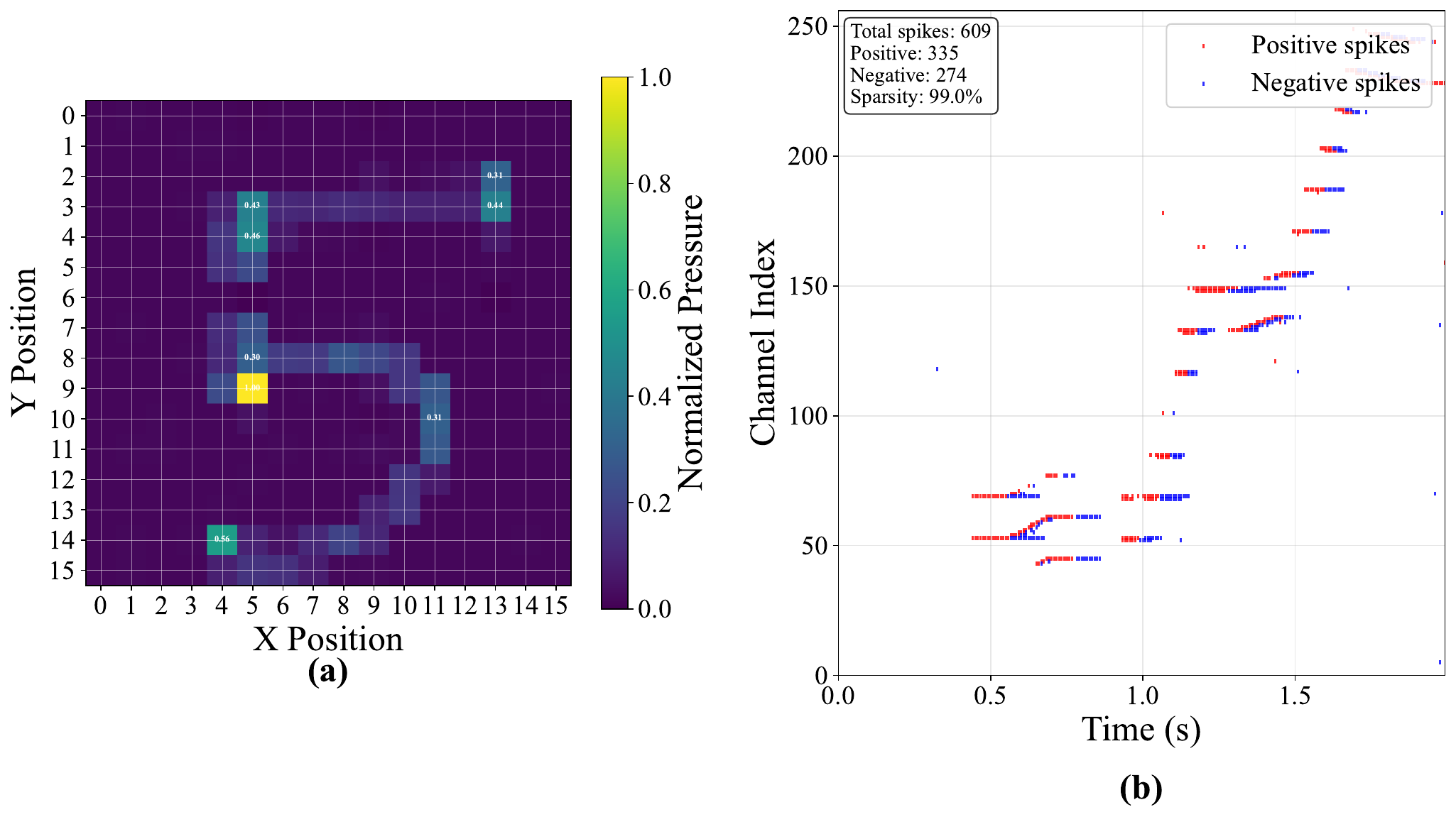} 
    \end{subfigure}
    \begin{subfigure}
        \centering
        \includegraphics[width=0.4\linewidth]{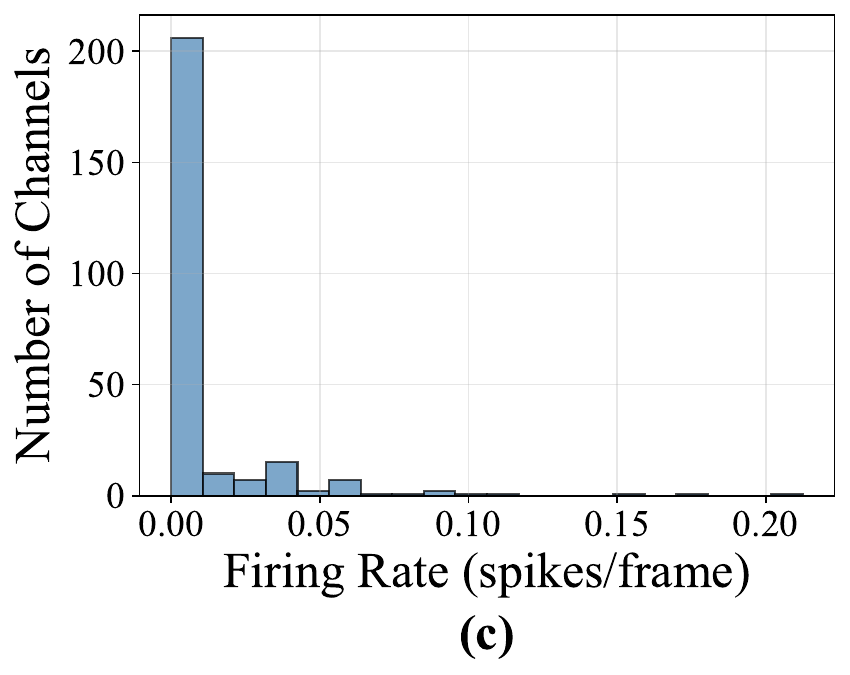}
    \end{subfigure}
    \begin{subfigure}
        \centering
        \includegraphics[width=0.4\linewidth]{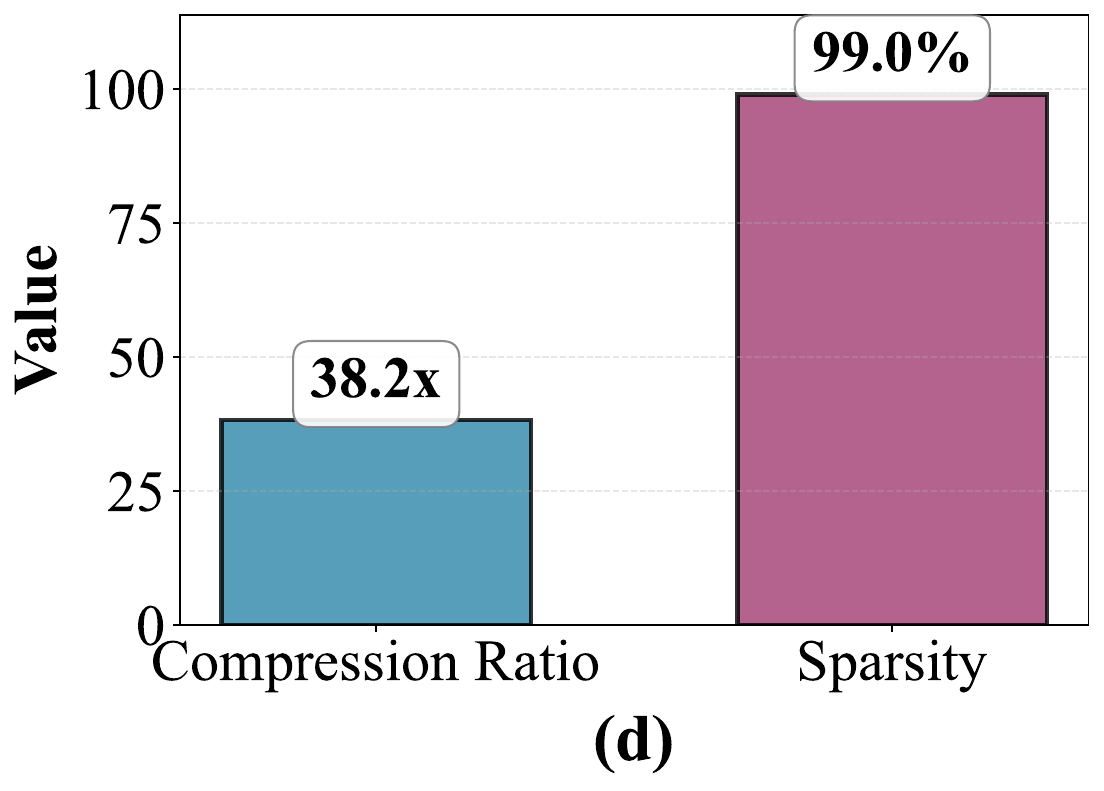}
    \end{subfigure}
    \caption{(a) Accumulated pressure map of digit "5", (b) Raster plot of spikes, (c) Firing rate histogram, (d) Efficiency in compression, storage and data sparsity}
    \label{fig_digit5}
\end{figure}

Figure \ref{fig_digit5}(a) presents the cumulative pressure map of a handwritten digit '5', illustrating the spatial pressure distribution during the writing process, while Fig. \ref{fig_digit5}(b) displays the corresponding raster plot of the compressed event sequence after delta modulation encoding, depicting the event-triggered state of each pixel based on threshold comparisons. At a compression ratio of $\Delta = 6$, the total number of pulses amounts to 609, including 335 positive and 274 negative pulses, yielding a sparsity of 99\%. 
Figure \ref{fig_digit5}(c) quantifies the event encoding statistics through a firing rate histogram, revealing a sparse distribution: approximately 200 pixels remain inactive, while about 50 highly active pixels reach firing rates up to 0.2 spikes per frame. To facilitate broader adoption in neuromorphic computing, the event data are packaged in AER format, incorporating pixel address, timestamp, and polarity. As demonstrated in Fig. \ref{fig_digit5}(d), this representation achieves a compression ratio of approximately 38.4× relative to raw data, resulting in 96\% storage savings and 99\% data sparsity.

\subsection{Recognition of Handwritten Digits}
The 760 samples were randomly split 4:1 into training and test sets for validation. For handwritten digit recognition, we employ a Conv-SNN comprising two convolutional layers followed by a fully connected layer, with a 16×16 multi-channel spike tensor as input.

\begin{table}[b]
\centering
\caption{Characteristic Comparison of Neuron Network}  
\label{tab:spd-compare}

\begin{tabular}{|c|c|c|c|}
\hline
Factors                                                                                        & CNN     & SNN    & \textbf{Conv-SNN(ours)} \\ \hline
Input Data Size&
  \begin{tabular}[c]{@{}c@{}}16×16\end{tabular} &
  \begin{tabular}[c]{@{}c@{}}240×16×16\end{tabular} &
  \begin{tabular}[c]{@{}c@{}}\textbf{240×16×16}\end{tabular} \\ \hline
\begin{tabular}[c]{@{}c@{}}Weight Memory \\ (bytes)\end{tabular}                               & 285,760 & 44,650 & \textbf{44,650}  \\ \hline
\begin{tabular}[c]{@{}c@{}}Average sparsity in \\ forward path\end{tabular}                    & 0.1315  & 0.9913 & \textbf{0.9997} \\ \hline
\begin{tabular}[c]{@{}c@{}}Forward- propagation\\ computational cost\\ (\# kMACs)\end{tabular} & 420.85  & 72.89   & \textbf{273.60 }    \\ \hline
Accuracy                                                                                       & 0.89  & 0.8618 & \textbf{0.9211}  \\ \hline
\end{tabular}
\label{Table}
\end{table}

To further underscore the advantages of Conv-SNN over SNN and CNNs, this study compare weight-deployment memory, data-path sparsity, forward-pass computational cost, and classification accuracy (Table~\ref{Table}). For the CNN, inputs are pressure maps reconstructed from 240 pressure frames; in contrast, spike inputs are derived directly from raw ADC signals via delta modulation, yielding ternary spikes ${-1,0,+1}$.

To ensure fairness, all architectures were matched for parameter count and convolutional hyperparameters, and LIF settings were identical for the SNN and Conv-SNN. Relative to CNNs, spike-based SNNs/Conv-SNNs exhibit greater robustness to weight quantization, permitting required weight memory reduced by 6.4×. Moreover, spike inputs are markedly sparser than raw adc data, enabling zero-skipping during inference. Thereby compared with CNN, our work substantially reduces MACs and overall forward-pass cost by 1.53×. The forward-pass cost in the Table \ref{Table} is determined by the required MAC operations per layer and the input-data sparsity of each convolutional and fully connected layers; Compared with CNN, our sparsity improvement is 7.6× based on the practical test set. In terms of classification accuracy, the accuracy of Conv-SNN significantly outperforms the plain SNN's accuracy by 6\%, indicating that convolutional kernels effectively extract spatial features in spike-based networks. The Conv-SNN also surpasses the CNN, which we attribute to its greater robustness to input noise under spike-based encoding. From the perspectives of computational cost and weight storage, spike-based neural networks are highly amenable to deployment on edge devices.

\begin{figure}[t]
    \centering
    \includegraphics[width=0.9\linewidth]{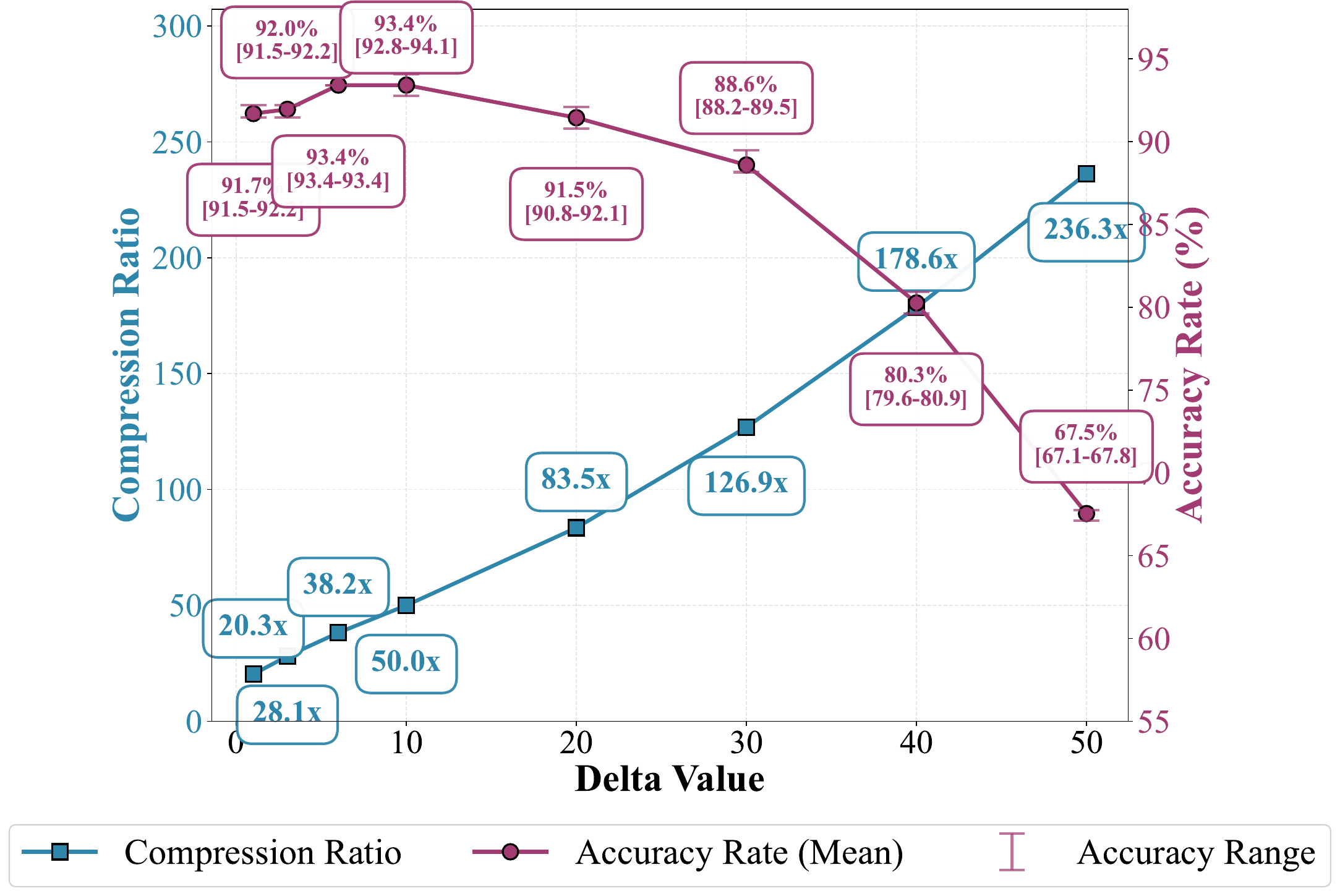}
    \caption{Trade-off between Compression Ratio (compared with raw data) and Classification Accuracy under Different Delta Values}
    \label{fig_com2}
\end{figure}

Additionally, this research studies the relationship between the input data compression ratio and the accuracy rate. As shown in Fig. \ref{fig_com2}, raising the delta threshold initially confers a denoising effect and thus get a Conv-SNN accuracy improvement by 1.7\%. Beyond this range, further compression discards more information during spike encoding, gradually degrading performance. At the optimal delta value with accuracy of 93.4\%, compared with CNN based the raw ADC data, the spiked based data compression ratio is 38.2×.

\section{Conclusion}
This paper presents a co-designed neuromorphic tactile system, integrating an event-driven binary scan strategy with Conv-SNN on FPGA. Our approach achieves substantial efficiency gains: the scanning strategy reduces scan counts by 12.8×, increase the DR by 28.4× and a 38.2× compression ratio, achieve 99\% data sparsity over conventional datasets. The Conv-SNN classifier efficiently processes these sparse event streams, requires only 65\% of the computation and 15.6\% of the weight storage relative to a CNN, all while maintaining a high accuracy of 92.11\%. Future work will focus on scaling the system for more complex tasks and real-world robotic deployment.

\end{document}